\documentclass[a4paper,conference]{IEEEtran}
\usepackage{cite}
\usepackage{hyperref}

\ifCLASSINFOpdf
    \usepackage[pdftex]{graphicx}
\else
    \usepackage[dvips]{graphicx}
\fi
\usepackage{amsmath}
\usepackage{amssymb}
\usepackage{array}
\usepackage{multicol}
\usepackage{multirow}
\usepackage{tabu}
\usepackage[dvipsnames]{xcolor}
\usepackage{url}

\begin{document}
%
\title{Portmanteauing Features for Scene Text Recognition}

\author{\IEEEauthorblockN{Yew Lee Tan*, Ernest Yu Kai Chew*\\ and Adams Wai-Kin Kong}
\IEEEauthorblockA{School of Computer Science \\and Engineering\\
Nanyang Technological University (NTU)\\
Singapore}
\and

\IEEEauthorblockN{Jung-Jae Kim\\ and Joo Hwee Lim}
\IEEEauthorblockA{Institute for Infocomm Research\\
Agency for Science, Technology and Research\\
Singapore}}


%


\maketitle

\begin{abstract}
Scene text images have different shapes and are subjected to various distortions, e.g. perspective distortions. To handle these challenges, the state-of-the-art methods rely on a rectification network, which is connected to the text recognition network. They form a linear pipeline which uses text rectification on all input images, even for images that can be recognized without it. Undoubtedly, the rectification network improves the overall text recognition performance. However, in some cases, the rectification network generates unnecessary distortions on images, resulting in incorrect predictions in images that would have otherwise been correct without it. In order to alleviate the unnecessary distortions, the portmanteauing of features is proposed. The portmanteau feature, inspired by the portmanteau word, is a feature containing information from both the original text image and the rectified image. To generate the portmanteau feature, a non-linear input pipeline with a block matrix initialization is presented. In this work, the transformer is chosen as the recognition network due to its utilization of attention and inherent parallelism, which can effectively handle the portmanteau feature. The proposed method is examined on 6 benchmarks and compared with 13 state-of-the-art methods. The experimental results show that the proposed method outperforms the state-of-the-art methods on various of the benchmarks.
\end{abstract}


%
\IEEEpeerreviewmaketitle

\section{Introduction}
\label{sec:introduction}
\let\thefootnote\relax\footnotetext{* these authors contributed equally to this work}
Scene text recognition (STR) is the task of reading text from natural scenes. The semantic property of text in images embodies additional information that is useful for many practical applications such as intelligent document processing, street signs reading, and product recognition \cite{long2021scene}. Because of these important applications, the field has drawn a great amount of attention from researchers and practitioners resulting in the emergence of various robust reading competitions \cite{lucas2005icdar, karatzas2015icdar, karatzas2013icdar, gomez2017icdar2017, rigaud2019icdar}. STR is challenging because the images are acquired in a generally unstructured and uncontrolled setting. This, in turn, introduces more complexity and variability than traditional optical character recognition (OCR) tasks \cite{chen2020text, zhu2016scene, long2021scene}. 

An approach in addressing such issues is to employ a rectification network to rectify perspective distorted and curved text before the text recognition process. Yang et al. \cite{yang2017learning} proposed the use of an auxiliary dense character detection scheme and an alignment loss to improve character localization in order to tackle the issues. In contrast, conceptually simpler methods which do not require explicit character detection are also introduced to address the problems \cite{shi2018aster, LUO2019109}.

\begin{figure}[ht!]
\begin{center}
\includegraphics[width=0.7\linewidth]{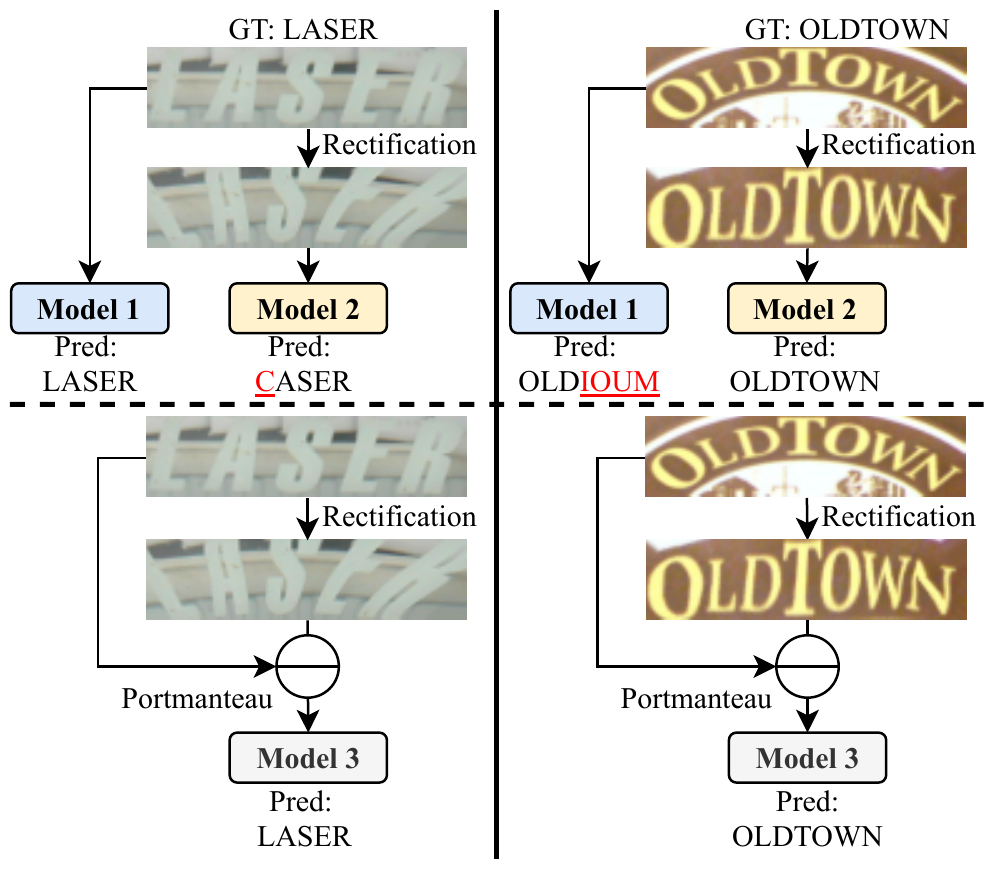} 
\end{center}
   \caption{The prediction results for multiple models from the experiments conducted. Model 1 uses the original image as input while model 2 utilizes a rectification module which rectifies the image before the recognition task. Model 3 uses both the rectified image as well as the original image as the inputs.}
\label{fig:intro_img1}
\end{figure}

Undoubtedly, the usage of rectification network has improved the overall performance of STR. However, the recognition network which comes after the rectification network may experience some form of penalty if, for example, suboptimal rectifications were being performed on regular text images. In other words, the recognition network will suffer from errors propagated from the rectification network, due to its linear input pipeline. An example is illustrated in Figure~\ref{fig:intro_img1} where the rectification of  ``LASER'' resulted in the loss of information and eventuated an incorrect prediction of ``CASER'' by model $2$.

To mitigate the potential problems that may arise from using the rectification network, this study proposes the \emph{portmanteau feature} which utilizes both the original and rectified images, together with the block matrix initialization, to form a non-linear pipeline. Like a portmanteau word whose meaning is derived from a blending of two or more morphemes (e.g. motel, brunch, smog), the portmanteau feature approach can better preserve the feature of the original image, which lightens the impact of unfavourable rectification, while incorporating the feature of the rectified image. This is shown in Figure~\ref{fig:intro_img1} where the proposed approach (model $3$) is able to generate the correct predictions in both cases as opposed to the model 1 which failed to predict ``OLDTOWN'' and model 2 which failed to predict ``LASER''.  

Other than using a rectification network to solve the irregularities of text images, an attention model is also often employed \cite{lu2019master,yu2020towards} due to its successes in both computer vision and natural language processing \cite{anderson2018bottom, vaswani2017attention}. The attention models can be used in tandem with a rectification network as it may be able to encode global context more effectively and hence, improve the overall performance of STR \cite{LUO2019109,hu2020gtc,lu2019master,li2019show}. 

Although the transformer model (which utilizes attention) \cite{vaswani2017attention} was conceptualized for natural language processing (NLP) tasks, it is becoming increasingly popular to be incorporated into vision related tasks due to its capabilities \cite{dosovitskiy2020image, bartz2019kiss, lu2019master}. However, the span of the transformer attention is limited to one axis of direction as NLP tasks mainly involve a sequence of information (or words). As an image contains information in both directions, Dosovitskiy et al. \cite{dosovitskiy2020image} split up the image into patches and arranged them into one sequence as demonstrated in their vision transformer (ViT). However, such method will incur a large amount of computational overhead.
Therefore, this work introduces the dual-axes vision transformer (DaViT) which is adapted from the Axial Transformer \cite{ho2019axial} in order to implicitly attend to features in two axes of direction.

The contribution of this work is as summarized: (1) portmanteau feature containing information from both original and rectified images, lightening the impact of unfavorable rectification,  and (2) the proposed network surpasses current state-of-the-art methods. 

The rest of the paper is organized as follows. Section~\ref{sec:relatedwork} explores related works. Section~\ref{sec:approach} discusses the approach and methodology. Section~\ref{sec:experiments} reports the experimental results on six scene text benchmarks, and Section~\ref{sec:conclusion} concludes this study.

\section{Related Work}
\label{sec:relatedwork}
Irregular texts present major challenges to STR, and so, Shi et al. and various other works \cite{shi2016end,shi2018aster,hu2020gtc,zhan2019esir} employed geometric transformation using thin plate splines (TPS) with spatial transformer network (STN). Subsequently, promising results from STN spurred several prominent works in text rectification. Luo et al. \cite{LUO2019109} developed a rectification method, called MORN, free from geometric constraints with the direct prediction of offset maps. Zhan and Lu \cite{zhan2019esir} proposed a STN based rectification network by iteratively rectifying the images, with controls points derived from a polynomial curve and line segments. Yang et al. \cite{yang2019symmetry} proposed a rectification module that includes prediction of text center line as well as the various local attributes, all derived from the ground-truth bounding box labels.

Attention mechanism is often utilized in tandem with rectification network in order to enhance the overall performance in STR. Lee and Osindero \cite{lee2016recursive} proposed the use of attention modeling which is conditioned upon the image feature from the CNN as well as the output of the RNN. Wang et al. \cite{wang2018memory} noticed that various STR works did not make full use of the alignment history and therefore, proposed the usage of a coverage vector to make adjustment for future attention.

Yu et al. \cite{yu2020towards} adopted the transformer architecture and introduced a parallel visual attention and a global semantic reasoning module for STR. The network employs a multi-way parallel transmission in order to make use of global semantic information more effectively. Bartz et al. \cite{bartz2019kiss} proposed the use of the vanilla transformer in their text recognition network which takes in regions of interest from their localization network.

\section{Approach}
\label{sec:approach}
\begin{figure*}[ht]
\begin{center}
\includegraphics[width=.69\linewidth]{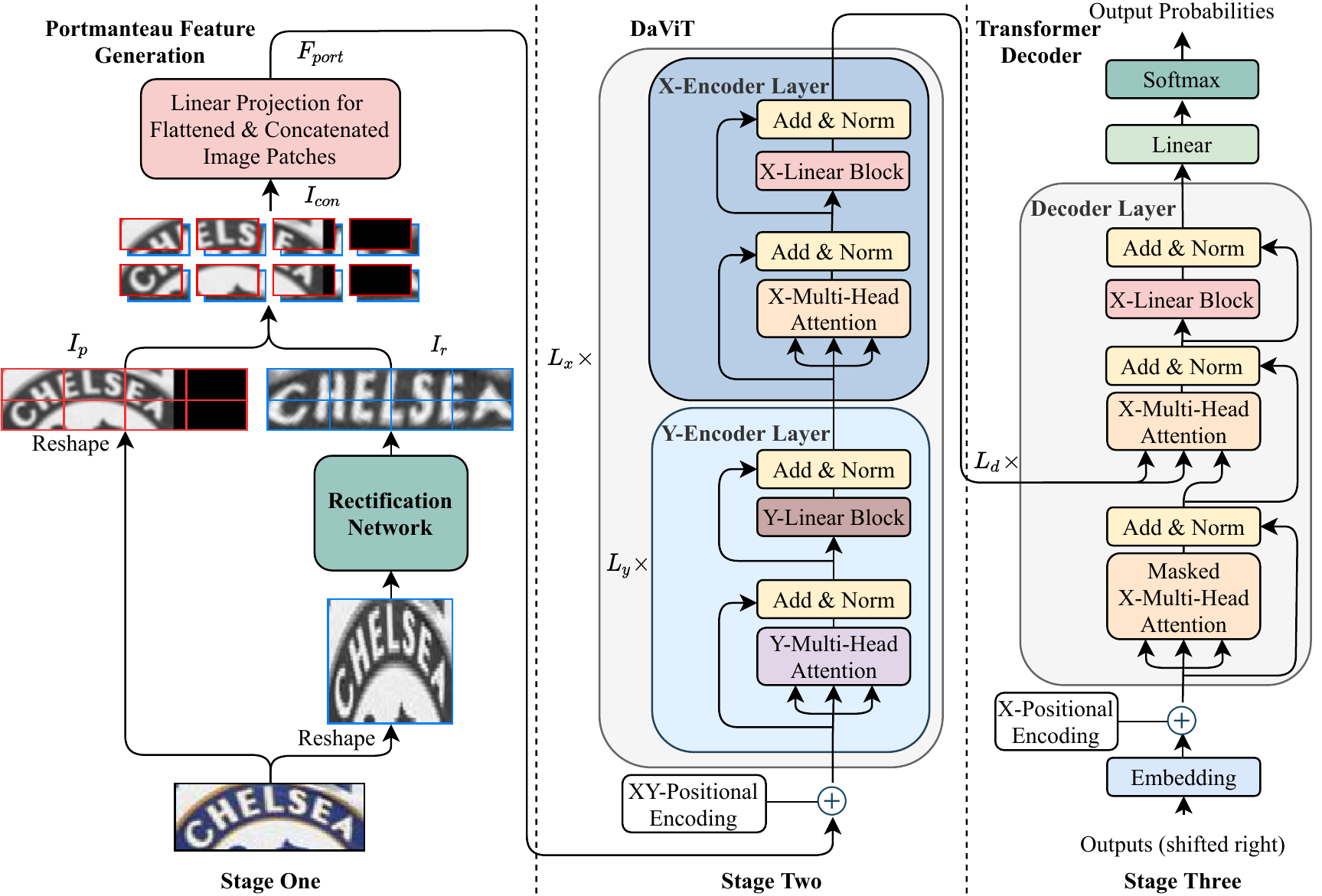}
\end{center}
\caption{Overview of the model, illustrating the three stages.}
\label{fig:model}
\end{figure*}

The proposed network can be broken down into three stages as shown in Figure~\ref{fig:model}. The first stage is the portmanteau feature generation where an input image goes through two different reshaping schemes. One of the schemes reshapes the image into a fixed height with varying width which will be padded to produce a padded image, $I_{p}$. The other scheme reshapes the image to a fixed height and fixed width followed by a rectification network to produce a rectified image, $I_r$. $I_r$ and $I_{p}$ then form a concatenated image, $I_{con}$, which goes through a linear projection block and produces the pormanteau feature, $F_{port}$. The linear projection block utilizes the block matrix initialization which allows the portmanteau feature to retain more individual information from the features of the rectified and padded images.

The second stage is the feature extraction stage that learns representation of the pormanteau feature. A variant of the ViT and Axial Transformer, called the DaViT, is proposed as the basic building block for this stage. By employing the transformer, the network will be able to achieve parallelism in the processing of the portmanteau feature. Furthermore, the DaViT is computationally less intensive than the ViT and Axial Transformer. The feature that transverses through the second stage will also be denoted as $F_{port}$.

The final stage of the network is the sequence modelling stage which incorporates contextual cues from the sequence of embeddings, as well as the output from the encoder in order to predict the next character. The stage is made up of the vanilla transformer decoder and the softmax function is used to predict the next character.

\subsection{Rectification}
Two types of rectification networks are adopted in this study separately, STN and MORN \cite{LUO2019109}. The TPS of STN makes geometric adjustments to straighten text, while MORN specializes in correcting complex distortion without applying geometric constraints.  

In this work, both STN and MORN used are pre-trained. The pre-trained STN follows ASTER STN \cite{shi2018aster}, but instead of predicting the control points directly, it predicts a $4$th order Legendre polynomial and $10$ line segments to generate the control points, similar to ESIR \cite{zhan2019esir}. 
Further details regarding the STN are available in the supplementary material\footnote{\hyperlink{https://github.com/chewyukai/portmanteauing-features}{https://github.com/chewyukai/portmanteauing-features}}. The SRNet\footnote{\hyperlink{https://github.com/youdao-ai/SRNet}{https://github.com/youdao-ai/SRNet}} framework used to train the STN, and the pre-trained MORN \footnote{\hyperlink{https://github.com/Canjie-Luo/MORAN_v2}{https://github.com/Canjie-Luo/MORAN\_v2}} are available on GitHub. 

\subsection{Portmanteauing Features}
\label{Portmanteauing}
A portmanteau word is a blending of two or more words such that the word expresses the combined meaning of its constituents. Like a portmanteau word, a concatenated connection is employed in the first stage of the framework in order to retain the information that exists in the padded image and the rectified image, as illustrated in Figure~\ref{fig:model}. 

\subsubsection{Concatenation}
After the reshape and rectification, $I_p \in \mathbb{R}^{H\times W \times 1}$ and $I_r \in \mathbb{R}^{H\times W\times 1}$ are concatenated to form the image $I_{con}= [I_{p}, I_{r}]$, where $I_{con} \in \mathbb{R}^{H \times W \times 2}$ and $H$, $W$ represent the height and width of the image. It is then split up into patches of $P_h \times P_w \times 2$ which is followed by a flattening, resulting in $I_{con} \in \mathbb{R}^{N_x \times N_y \times 2P_wP_h}$, where $N_x=\frac{W}{P_w}$ and $N_y=\frac{H}{P_h}$. $I_{con}$ will then enter a linear projection block (described in Section~\ref{sec:LPB}), which utilizes the block matrix initialization (described in Section~\ref{sec:BlockInit}), and be projected into a portmanteau feature, $F_{port}$. In the experiments, $P_w=2$ and $P_h=4$. 

\subsubsection{Block Matrix Initialization}
\label{sec:BlockInit}
Through the utilization of the transformer's intrinsic property (i.e. the multi-head attention module), the network is able to achieve parallelism in processing the portmanteau feature as if it is going through two separate transformers at the same time. 

The block matrix initialization (BMI) scheme is proposed for the parameters in the linear layers in order to regulate the interaction between them. By initializing the parameters as such, it is assumed that the optimal solution exists within the premise of limited interaction and the network will be able to learn to facilitate the exchange of information with restraint.

The BMI enables $I_{con}$ and $F_{port}$ to start off as though $I_{con}$ is going through two separate linear projection blocks followed by $F_{port}=[F_p, F_r]$ going through two separate transformers. For an arbitrary input $ X = [X_{1}, X_{2}]$ where $X_1,X_2 \in \mathbb{R}^{N \times D_{in}}$, and $X \in \mathbb{R}^{N \times 2D_{in}}$. 
The transformation matrix in the linear layer, $A \in \mathbb{R}^{2D_{in} \times 2D_{out}}$, can be broken down into a block matrix as such:
\begin{equation}
    A =
    \begin{bmatrix}
    A_{11} & A_{12} \\
    A_{21} & A_{22}
    \end{bmatrix}
    , A_{ij} \in \mathbb{R}^{D_{in} \times D_{out}}
\end{equation}
By having $A_{12}$ and $A_{21}$ as zero matrices, the linear transformation on $X$ is as though both $X_1$ and $X_2$ are going through their individual linear transformation with no interaction between them as follows:
\begin{equation}
    XA = 
    \begin{bmatrix}
    X_{1}A_{11} & X_{2}A_{22} 
    \end{bmatrix}
\end{equation}
The two off-diagonal block matrices in each of the linear layers of the encoder are initialized as zero matrices while the diagonal matrices are initialized with the Xavier initialization \cite{glorot2010understanding}.

\subsubsection{Linear Projection}
\label{sec:LPB}
The linear projection (LP) consists of two linear transformations as shown in Equation~\ref{eqn:LP}.
\begin{equation}
    LP(I_{con}) = (I_{con}A_1+b_1)A_2+b_2
    \label{eqn:LP}
\end{equation}
The output of the LP is the portmanteau feature $F_{port}=LP(I_{con})$. 
For the experiments, the input dimension is $2P_wP_h=16$, the output dimension $D_y$ which represents the dimensionality of the y-encoder layer is $96$, and the inner dimension $D_{LP}$ is $2048$.

\subsection{Dual-Axes Vision Transformer Encoder}
\label{DaViT}
Scene text images can be densely-packed with characters. Hence, the size of image patches is required to be small enough such that the embedding of every character in the image can be learned. However, the reduction in image patch dimension increases the sequence length.
As the complexity of the transformer self-attention layer is $O(N^2D)$ where $D$ is the dimension of the representation, using such a method would be computationally intensive.

In order to solve the above-mentioned issue, an adaptation of the Axial Transformer \cite{ho2019axial}, DaViT encoder is proposed. Instead of an explicit two-dimensional attention of the ViT, Axial Transformer alternates between vertical and horizontal attentions, thereby allowing the model to attend in two axes of direction. The range of patches that the attention can explicitly access will be more limited as compared to ViT in exchange for the reduction in computational overhead. The DaViT further reduces the computational complexity as shown in Table~\ref{tab:complex}.

\begin{table}[ht]
\begin{center}
\caption{Complexity of ViT and DaViT self-attention layer.
$L_y$ is the number of layers for the y-axis encoder.}
\label{tab:complex}
\begin{tabular}{|c|c|}
\hline
\bf{Attention Type} & \bf{Complexity per Layer} \\
\hline\hline
ViT  & $O(N_x^2\cdot N_y^2\cdot D_x)$ \\
\hline
Axial Transformer & $O(N_x^2\cdot D_x \cdot N_y + N_y^2\cdot D_x \cdot N_x )$ \\
\hline
DaViT & $O(N_x^2\cdot D_x + N_y\cdot N_x \cdot D_x \cdot L_y )$ \\
\hline
\end{tabular}
\end{center}


\end{table} 

The transformer encoder layer that has an attention span across the x-axis is called the x-encoder layer while the one that has its attention span across the y-axis will be called the y-encoder layer as illustrated in the stage two of Figure~\ref{fig:model}. The y-encoder layer follows its respective architecture in the Axial Transformer where the $N_x=\frac{W}{P_w}$ columns of features will go through the attention mechanism. The feature which transverses through the y-encoder layers with the shape $N_x \times N_y \times D_y$ will have the $N_x$ technically treated as a ``mini-batch'' size with $N_y=\frac{H}{P_h}$ representing the sequence length.

For the x-encoder layers, $F_{port}$ will be reshaped into $N_x \times D_x$ where the concatenation of $D_y$ through $N_y$ forms $D_x$ ($D_x=D_y \times N_y$) and $N_x$ will represent the sequence length. This is in contrast with the Axial Transformer where the input to corresponding encoder layer would be of the dimension $N_y \times N_x \times D_x$ and the features dimension is the same for both directions ($D_x=D_y$). 
Specifically for each y-encoder layer, the complexity per layer for DaViT is $N_y^2\cdot D_y=N_y\cdot N_x \cdot D_x$. Conversely, the counterpart in Axial Transformer has a complexity per layer of $N_y^2\cdot D_x \cdot N_x$. For the x-encoder layer, the complexity per layer for DaViT is $N_x^2\cdot D_x$ whereas the complexity for the Axial Transformer is $N_x^2\cdot D_x \cdot N_y$.  In the experiments, the number of x-encoder layers $L_x=8$, the number of y-encoder layers $L_y=2$, $N_x=64$ and $N_y=8$.

\subsubsection{Multi-Head Attention}
Since DaViT performs attention sequentially across both the x and y directions, the model consists of both x and y multi-head attention with each having an attention spanning across the x-axis and y-axis respectively. According to the transformer \cite{vaswani2017attention}, the multi-head attention (MHA) is as follows:
\begin{equation}
    SDPA(Q,K,V) = softmax(\frac{QK^{\intercal}}{\sqrt{D/h}})V
\end{equation}
\begin{equation}
    MHA(Q,K,V) = Concat(Att_0,...,Att_{h-1})A_M
\end{equation}
where $Att_i = SDPA(QA_{Q_i},KA_{K_i},VA_{V_i})$, $i$ represents the index of the head, and the transformation matrices $A_{Q_i},A_{K_i},A_{V_i} \in \mathbb{R}^{D \times \frac{D}{h}}$. For the y-encoder layers, $h_y=2$ and for the x-encoder layers, $h_x=16$.

\subsubsection{Linear Block}
The linear block (LB) in the network applies two linear layers to each position in the sequence identically, followed by a rectified linear unit (ReLU):
\begin{equation}
 LB(F_{port}) = max(0, F_{port}A^{\prime}_1+b^{\prime}_1)A^{\prime}_2+b^{\prime}_2   
\end{equation}
where $k\in \{x,y\}$, $A^{\prime}_1 \in \mathbb{R}^{D_k \times D_{LB_k}}$, and $A^{\prime}_2 \in \mathbb{R}^{D_{LB_k} \times D_k}$. X-linear blocks apply the linear transformations to the sequence along the x-direction whereas the y-linear blocks apply to the y-direction. For the x-linear blocks, the input and output are of the dimension $D_x=768$, and the inner-layer has the dimensionality of $D_{LB_x}=2048$. The input and output to the y-linear blocks have the dimensionality of $D_y=96$ with $D_{LB_y}=512$.

\subsection{Transformer Decoder}
\label{sec:decoder}
All the building blocks in the decoder are similar to that of DaViT as can be seen on Figure~\ref{fig:model}. However the masked multi-head attention applies a mask which prevents the attention of subsequent positions (i.e. looking into the future). The decoder is accompanied by a softmax classfier which predicts the probability distribution over the character classes. For the experiments, the number of decoder layers, $L_d=4$.

\section{Experiments}
\label{sec:experiments}
\subsection{Implementation Details}

\label{ImplementationDetails}
The network was implemented using PyTorch and trained using the ADAM optimizer with a base learning rate of $0.02$, betas of $(0.9, 0.98)$, and eps of $1e^{-9}$. The warmup duration was set to $6000$ iterations with a decaying rate of $\min({steps^{-0.5}}, steps \times  warmup^{-1.5})$. The model was trained on 8× NVIDIA RTX6000 24GB, with a batch size of $768$ for $300,000$ iterations.

The proposed network was trained with three synthetic datasets: \emph{MJSynth} (MJ) \cite{jaderberg2014synthetic} containing $9$ million samples, \emph{SynthText} (ST) \cite{Gupta16} containing 8 million images, and \emph{SynthAdd} (SA) \cite{li2019show} containing 1.6 million images used to compensate for the lack of punctuation in \emph{MJSynth} and \emph{SynthText}. 

\subsubsection{Scene Text Benchmarks}
The proposed network was evaluated on six scene text benchmarks. \emph{IIIT 5K-Words} (IIIT) \cite{mishra2012scene} contains $3000$ test images.\emph{ICDAR 2013} (IC13) \cite{karatzas2013icdar} contains $1015$ images for testing where samples that contain non-alphanumeric characters, or have less than three characters were discarded.\emph{ICDAR 2015} (IC15) \cite{karatzas2015icdar} contains $2077$ test images.\emph{Street View Text} (SVT) \cite{wang2011end} consists of $647$ testing images.\emph{Street View Text-Perspective} (SVTP) \cite{phan2013recognizing} has $645$ testing images.\emph{CUTE80} \cite{risnumawan2014robust} (CT) contains $288$ testing images.



\begin{table*}[ht!]

\small
\caption{Text recognition accuracy (in percentages) on six scene text benchmarks. For each benchmark, the result shown in {\bfseries bold}. represents the best performance, while \underline{underline} represents the second best. ST, MJ, and SA denote the synthetic datasets of SynthText, MJSynth, and SynthAdd respectively. R denotes the use of real datasets in training or fine-tuning. Port$_{MORN}$ and Port$_{STN}$ represent portmanteau feature generation using MORN and STN, respectively.}
\begin{center}
\begin{tabular}{|c|c|c|c|c|c|c|c|}
\hline
\multirow{3}{*}{\bf{Method}} & \bf{Train} & \multicolumn{3}{c|}{\bf{Regular Text}} & \multicolumn{3}{c|}{\bf{Irregular Text}} \\ \cline{3-8}
& \bf{Datasets} & IIIT & IC13 & SVT & IC15 & SVTP & CT \\ \cline{3-8}
& & 3000 & 1015 & 647 & 2077 & 645 & 288 \\
\hline\hline
Luo et al. \cite{LUO2019109} & ST+MJ & 91.2 & 92.4 & 88.3 & 68.8 & 76.1 & 77.4 \\ 
Yang et al. \cite{yang2019symmetry} & ST+MJ & 94.4 & 93.9 & 88.9 & 78.7 & 80.8 & 87.5 \\
Zhan and Lu \cite{zhan2019esir} & ST+MJ & 93.3 & 91.3 & 90.2 & 76.9 & 79.6 & 83.3 \\
Wang et al. \cite{wang2020decoupled} & ST+MJ & 94.3 & 93.9 & 89.2 & 74.5 & 80.0 & 84.4 \\
Yu et al. \cite{yu2020towards} & ST+MJ & 94.8 & \underline{95.5} & 91.5 & -  & 85.1 & 87.8 \\
Zhang et al. \cite{zhang2020autostr} & ST+MJ & 94.7 & 94.2 & 90.9 & - & 81.7 & - \\ 
Qiao et al. \cite{qiao2020seed} & ST+MJ & 93.8 & 92.8 & 89.6 & 80.0 & 81.4 & 83.6 \\
Bartz et al. \cite{bartz2019kiss}& ST+MJ+SA & 94.6 & 93.1 & 89.2 & 74.2 & 83.1 & \bf{89.6} \\
\hline
Port$_{STN}$+DaViT & ST+MJ & 93.8 & 95.0 & \bf{94.3} & \underline{80.3}  & \bf{89.9} & \underline{88.5} \\
Port$_{MORN}$+DaViT & ST+MJ+SA & \bf{95.3} & \bf{96.0} & 91.0 & 79.9  & \underline{87.8} & 83.3 \\ 
Port$_{STN}$+DaViT & ST+MJ+SA & \underline{95.1} & 94.3 & \underline{92.3} & \bf{80.5}  & 87.6 & 86.8 \\
\hline
\hline
Yue et al. \cite{yue2020robustscanner} & ST+MJ+R & \underline{95.4} & \underline{94.1} & 89.3 & \underline{79.2} & 82.9 & \bf{92.4} \\
Wan et al. \cite{wan2020textscanner} & ST+MJ+R & \bf{95.7} & \bf{94.9} & \underline{92.7} & - & \underline{84.8} & \underline{91.6} \\
\hline
Port$_{STN}$+DaViT & ST+MJ+R & 95.0 & \underline{94.1} & \bf{94.1} & \bf{82.5}  & \bf{90.0} & 87.8 \\
\hline
\hline
Li et al. \cite{li2019show} & ST+MJ+SA+R & \underline{95.0} & 94.0 & 91.2 & 78.8 & \underline{86.4} & \underline{89.6} \\
Lu et al. \cite{lu2019master}& ST+MJ+SA+R & \underline{95.0} & \bf{95.3} & 90.6 & \underline{79.6} & 84.5 & 87.5\\
Hu et al. \cite{hu2020gtc} & ST+MJ+SA+R & \bf{95.8} & \underline{94.4} & \underline{92.9} & 79.5 & 85.7 & \bf{92.2}  \\
\hline
Port$_{STN}$+DaViT & ST+MJ+SA+R & \underline{95.0} & 93.6 & \bf{94.6} & \bf{82.7} & \bf{87.0} & 89.2 \\
\hline

\end{tabular}
\end{center}

\label{tab:result}
\end{table*}

\begin{table*}[t]

\small
\caption{Result of ablation study for portmanteau feature, in terms of text recognition accuracy (in percentages). The `(U)' in the method names represents resizing the images without padding.}
\label{tab:ablation-port}
\begin{center}
\begin{tabular}{|c|c|c|c|c|c|c|c|}
\hline
\multirow{3}{*}{\bf{Method}} & \multicolumn{3}{c|}{\bf{Regular Text}} & \multicolumn{3}{c|}{\bf{Irregular Text}} & \bf{Avg.} \\ \cline{2-7}
& IIIT & IC13 & SVT & IC15 & SVTP & CT & \bf{Acc}\\ \cline{2-8}
& 3000 & 1015 & 647 & 2077 & 645 & 288 & 7672\\
\hline\hline
Port$_{STN}$+DaViT & 95.1 & 94.3 & 92.3 & 80.5 & 87.6 & 86.8 & 89.9\\
Port$_{STN}$+DaViT (U) & 94.8 & 95.5 & 93.7 & 75.7 & 88.2 & 89.6 & 88.9\\
Port$_{STN}$+DaViT w/o BMI &
94.9 & 
94.9 & 
92.3 &
80.2 &
87.3 &
87.2 &
89.8
\\

Port$_{STN}$+DaViT w/o BMI (U) &
94.7 & 
96.4 & 
93.5 &
75.8 &
87.4 &
88.5 &
88.7
\\
STN+DaViT & 
95.1 & 
93.9& 
92.1& 
79.6& 
87.9& 
86.8 &
89.6
\\

DaViT & 
94.3& 
95.7& 
90.9& 
77.0& 
83.4& 
84.7&
88.2
\\

\cline{1-8}
\end{tabular}
\end{center}

\end{table*}

\subsection{Experimental Results}
The network recognizes $100$ token classes, including $10$ digits, $52$ case sensitive letters, $35$ punctuation characters, a start token, an end token, and a pad token. During evaluation, only $37$ classes which include $36$ case-insensitive alphanumeric characters and an end token were taken into consideration. The maximum output sequence length is $30$ and greedy decoding was used. No rotation strategy \cite{li2019show} was used in the experiments.

The network was evaluated on six benchmarks and the comparison results are given in Table \ref{tab:result}. The results show that the proposed network with MORN (Port$_{MORN}$+DaViT) outperforms the original MORAN \cite{LUO2019109} significantly for all the test datasets. Moreover, both Port$_{MORN}$+DaViT and Port$_{STN}$+DaViT which utilize a full Transformer encoder-decoder architecture outperforms KISS \cite{bartz2019kiss} and MASTER \cite{lu2019master}, where both networks are ResNet-Transformer hybrids. The result demonstrates the benefit of the transformer encoder over the ResNet. 

Furthermore, the proposed models achieved state-of-the-art result for the majority of the datasets with Port$_{STN}$+DaViT leading IC15 (containing 2077 images) by $0.9\%$ and SVT-P (containing 645 images) by $1.2\%$. Comparing the results of other works which included real datasets in the training or fine-tuning process, Port$_{STN}+DaViT$ achieved $82.7\%$ accuracy on IC15 and $94.6\%$ on SVT, outperforming the next best results from other works by $2.7\%$ and $1.7\%$ respectively. Although there is a drop in the accuracy for SVTP when real datasets are included, the result of $87.0\%$ still outperforms the next best result from otherworks by $0.6\%$.

\subsection{Ablation Studies}
\label{sec:ablation}
\subsubsection{Discussion on Portmanteauing Feature}




\begin{figure*}[ht!]
\begin{center}
   \includegraphics[width=0.8\linewidth]{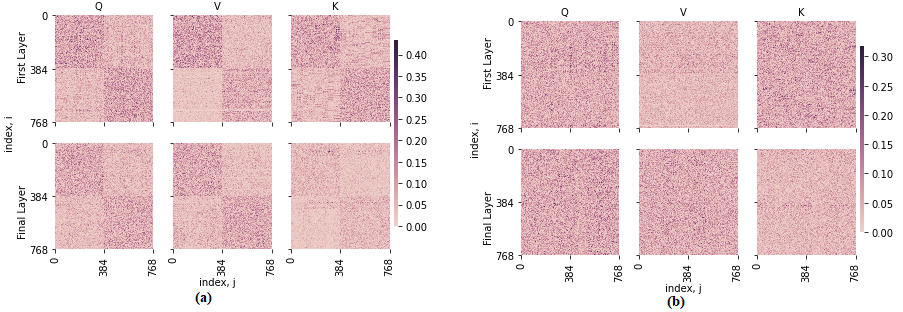}
\end{center}
   \caption{Heat maps of the Q, V, K linear layers parameters for the x-multi-head attention in the first and last encoder layers for (a) with BMI and (b) without BMI. The absolute values of the parameters are shown to illustrate with a better contrast.}  
\label{fig:portmanteau_heatmap}
\end{figure*}

Two additional networks were trained without the portmanteau feature in order to investigate its impact as follows: (1) the STN+DaViT which contains a rectification network, and (2) the DaViT with only $I_p$ as the input. The two networks were trained with the same hyperparameters as Port$_{STN}$+DaViT. 

\begin{figure}[ht]
\begin{center}
   \includegraphics[width=0.9\linewidth]{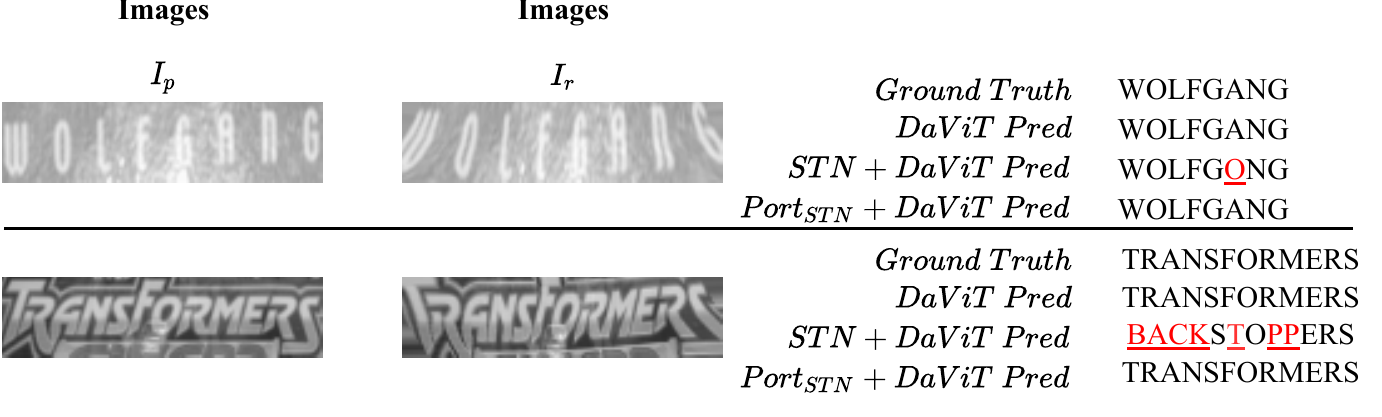}
\end{center}
   \caption{Examples from the test datasets on the adverse effect of STN on the predictions.}  
\label{fig:STN_adverse}
\end{figure}

Table~\ref{tab:ablation-port} shows the result of the experiments. To a certain extent, it demonstrates that the introduction of the STN causes an adverse impact to the text recognition capability of the model. Such effect is explicitly evident on the IC13 dataset as the accuracy dropped by $1.8\%$ when STN is added into DaViT although the average accuracy on the datasets increases. Figure~\ref{fig:STN_adverse} shows some examples from the test datasets on the adverse effect of the STN on the predictions and the alleviation of the effect by portmanteaing of features. More examples are provided in the supplementary material.

\begin{figure}[ht]
\begin{center}
   \includegraphics[width=0.9\linewidth]{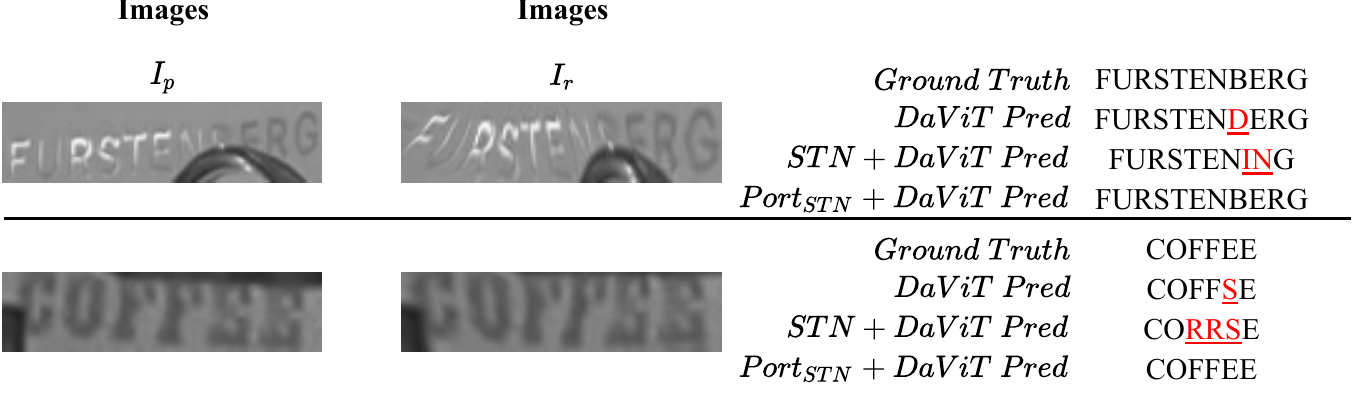}
\end{center}
   \caption{Examples from the test datasets.}  
\label{fig:STN_adverse2}
\end{figure}

Comparing Port$_{STN}$+DaViT with STN+DaViT and DaViT, where only one image was used, shows that the portmanteauing of feature generally improves the overall accuracy of the test datasets. The network, appears to be able to mitigate the adverse effect of STN as it outperforms STN+DaViT on four benchmarks including IC13 dataset. The incorporation of the features from $I_p$ and $I_r$ also allows the network to attain higher accuracy on SVT and IC15 than what could have been achieved individually. Some examples are shown in Figure~\ref{fig:STN_adverse2} where the Port$_{STN}$+DaViT is able to attain correct predictions for the text images where both DaVit and STN+DaViT result in wrong predictions.  More examples are provided in the supplementary material.


From Table~\ref{tab:ablation-port}, it appears that BMI slightly improves the average accuracy of the test datasets. It is interesting to note that the parameters of the linear layers in Port$_{STN}$+DaViT with BMI is able to retain the characteristic of the block matrix initialization after the training as seen on Figure ~\ref{fig:portmanteau_heatmap}. Such patterns can be observed in all linear layers where the block matrix initialization was used. This suggests that a more optimal solution exists where the interaction between the features of the two images are regulated by the BMI.

Furthermore, results from Table~\ref{tab:ablation-port} show that using the same reshaping scheme of unpadding for both images (i.e. method names denoted by ``U") results in worst overall performance. Although some datasets benefited from this, there is a drop of about 1\% in average accuracy across the 6 benchmarks as compared with their counterparts (without ``U").

\section{Conclusion}
\label{sec:conclusion}
In this paper, the adverse impact of rectification network was discussed. Although the rectification network improves overall text recognition accuracy, this study found that in some instances, rectification network can produce undesirable results. In order to address this issue, the portmanteauing of feature was proposed together with DaViT which can effectively handle the portmanteau feature. Experimental results show that the approach, to a certain extent, was able to lighten the adverse impact of rectification and also achieved the state-of-the-art results for most scene text recognition benchmarks.

\section{Acknowledgement}
This work is partially supported by NTU Internal Funding -- Accelerating Creativity and Excellence (NTU–ACE2020-03).

\bibliographystyle{IEEEtran}

%




\end{document}


%
\title{Supplementary Material for: Portmanteauing Features for Scene Text Recognition}

\author{\IEEEauthorblockN{Yew Lee Tan, Ernest Yu Kai Chew\\ and Adams Wai-Kin Kong}
\IEEEauthorblockA{School of Computer Science \\and Engineering\\
Nanyang Technological University\\
Singapore}
\and

\IEEEauthorblockN{Jung-Jae Kim\\ and Joo Hwee Lim}
\IEEEauthorblockA{Institute for Infocomm Research\\
Agency for Science, Technology and Research\\
Singapore}}


%


\maketitle


%
\IEEEpeerreviewmaketitle

\section{STN}
\label{app:STN}
\subsection{Input Protocols for Portmanteau Features}
To generate portmanteau features, the raw images were first converted into grayscale images, before being normalized via two transformation protocols. 

The first protocol resized the image to a fixed height of $32$ pixels with varying width to retain the aspect ratio. After this resizing, if the width of the image was shorter than $128$ pixels, zero padding would be applied. Otherwise, the width of the image would be further resized to $128$ pixels.  

The second protocol directly resized the input image to $300 \times 300$ pixels for STN and $32 \times 128$ pixels for MORN. The size of the output images from the two rectification networks is $32 \times 128$ pixels. Both protocols would yield grayscale images with a size of $32 \times 128$ pixels.  

The STN was trained with a batch size of $32$ for $150$ epochs, with a configuration of initial learning rate as $5e^{-4}$, the warm-up iteration was $10,000$, and the decay per iteration was set such that the final learning rate is one-fifth of its initial. The Adam optimizer used was configured with betas of $(0.9, 0.999)$, and an eps of $1e^{-8}$. Lastly, the STN dataset contains $80,000$ synthetic images with a $9$-to-$1$ train-validation split.

In addition, the STN also adopts a custom loss function, which is defined as:  
\begin{multline}
L_{STN} = \mid\mid \psi - \hat{\psi} \mid\mid +  \alpha\mid\mid \phi - \hat{\phi} \mid\mid \\+\beta\mid\mid \cos{\theta} - \cos{\hat{\theta}} \mid\mid +\gamma \mid\mid \sin{\theta} - \sin{\hat{\theta}} \mid\mid + \delta\mid\mid \xi - \hat{\xi} \mid\mid
\end{multline}
where $\mid\mid \bullet \mid\mid$ represents L1 norm, $\hat{\psi}$ is the estimated coefficients of the $4$th order Legendre polynomial, $\hat{\theta}$ is the estimated angles between the line segments and the tangents of the polynomial, $\hat{\phi}$ is the estimated x-coordinates of the intersection points between the line segments and the polynomial, and $\hat{\xi}$ is the length of the line segments. $\psi, \theta, \phi$, and $\xi$ are the corresponding ground truths and $\alpha, \beta, \gamma,$ and $\delta$ are hyperparameters. It should be highlighted that the area between the predicted and ground-truth polynomial is directly proportional to the difference between their Legendre coefficients. 




\subsection{Polynomial Scheme for STN}
The polynomial scheme could be dividing into six parts, as follows:
\begin{enumerate}

\setlength{\parskip}{0pt}
\item  extracting the key points from the ground-truth bounding boxes
\item  fitting the polynomial curve and lines, using the extracted key points
\item  determining the intersection points between the polynomial and its line segments
\item  interpolating the intersection points and the line angles to get the desired  number of line segments
\item  converting the coefficients of the fitted polynomial into its Legendre equivalence
\item combining all the variables into a single loss function
\end{enumerate}

\subsubsection*{Key points extraction}
The first step in the polynomial scheme was to resize the image into a square image. The height and width of the square image were considered to be in the range of -1 and 1. 

\begin{figure}[ht!]

\begin{center}
   \includegraphics[]{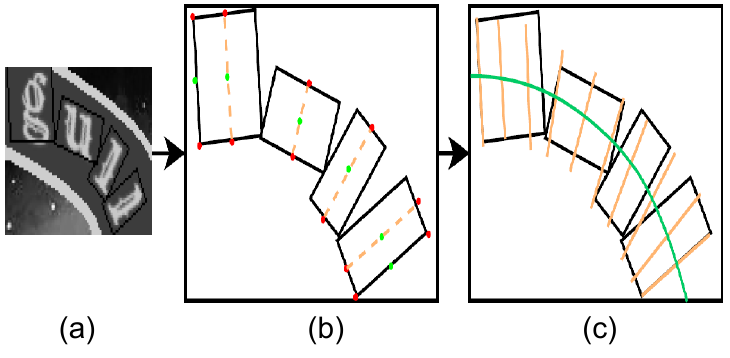}
\end{center}
\caption{Overview of the STN polynomial scheme, where the ground-truth bounding boxes were used to derive a 4th order polynomial with 10 lines segments.}  
\label{fig:STN-GT}
\end{figure}

Subsequently, the top-center, center and bottom-center of each bounding box, in addition to the top, center and bottom of the left-most and right-most corners, were extracted as key points for curve and line fitting. These key points were denoted as the green and red points in Figure~\hyperref[fig:STN-GT]{1b}.

\subsubsection*{Curve and line fitting}
After which, the center key points were used to fit a 4th order polynomial, while the top, center, and bottom points were used to fit ($n+2$) linear functions, where $n$ was the number of characters in the word. 

\subsubsection*{Intersection between polynomial and line}
Then, the intersection points between the polynomial and the $n+2$ line segments were determined and denoted as $\phi_{raw}$. The angles between the tangent lines and the line segments were also calculated and denoted as $\theta_{raw}$. In addition, the height of the tallest bounding box was computed and denoted as $\xi$. 

\subsubsection*{Line segments interpolation}
Next, $\phi_{raw}$ and $\theta_{raw}$ would be used as key reference points for rotational and translational interpolation to produce the desired number of line segments (from $n+2$) of length $\xi$. In this paper, the desired number of line segments was set to 10 (See Figure~\hyperref[fig:STN-GT]{1c}).

\subsubsection*{Legendre polynomial}

\begin{figure}[ht!]
\begin{center}
   \includegraphics[width=0.6\linewidth]{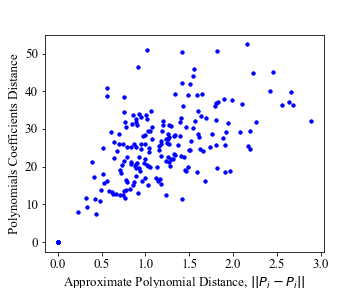}
   \includegraphics[width=0.6\linewidth]{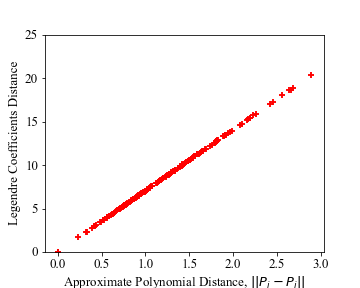}
\end{center}
\caption{The scatter plots of polynomial coefficients distance (top) and Legendre coefficients distance (bottom) of 400 polynomial pairs against $\lVert P_i - P_j \rVert$.} 
\label{fig:STN_PCvLC}
\end{figure}

Then, the coefficients of the polynomial would be converted to its Legendre equivalence. This is because Legendre polynomials form a complete orthogonal basis on $L^2$~[-1,1] ($L^2$~space) \cite{GUMEROV200439} and so, their coefficients may be better for measuring the similarity between two polynomials.

Figure~\ref{fig:STN_PCvLC} was constructed with 20 random polynomials $f_i(x)$, where $i \in [1,20]$. Each polynomial was paired with each other to form 400 polynomial pairs. $N=201$ evenly distributed points across the $x$-axis from $[-1, 1]$ were sampled from each polynomial and the $y$ values form a point, $P_i$, where $P_i \in \mathbb{R}^{N}$. The 400 polynomial pairs were represented as in the scatter plots, in terms of the distance, $\lVert P_i - P_j \rVert$, and the $\lVert coef(f_i) - coef(f_j) \rVert$, where $i,j \in [1,20]$ and $coef(\dot)$ returns the coefficients of its polynomial function. 

Although the polynomial coefficient distances seemly correlate with the distance $\lVert P_i - P_j \rVert$ (top of Figure~\ref{fig:STN_PCvLC}), they did not demonstrate proportionality like its Legendre equivalence (bottom of Figure~\ref{fig:STN_PCvLC}), which was the reason Legendre coefficients were used in this work instead.

\subsubsection*{Loss function}
\begin{multline}
L_{STN} = \mid\mid \psi - \hat{\psi} \mid\mid +  \alpha\mid\mid \phi - \hat{\phi} \mid\mid \\+\beta\mid\mid \cos{\theta} - \cos{\hat{\theta}} \mid\mid +\gamma \mid\mid \sin{\theta} - \sin{\hat{\theta}} \mid\mid + \delta\mid\mid \xi - \hat{\xi} \mid\mid
\end{multline}
where $\mid\mid \bullet \mid\mid$ represents L1 norm, $\hat{\psi}$ is the estimated coefficients of the $4$th order Legendre polynomial, $\hat{\theta}$ is the estimated angles between the line segments and the tangents of the polynomial, $\hat{\phi}$ is the estimated x-coordinates of the intersection points between the line segments and the polynomial, and $\hat{\xi}$ is the length of the line segments. $\psi, \theta, \phi$, and $\xi$ are the corresponding ground truths and $\alpha, \beta, \gamma,$ and $\delta$ are hyperparameters.

\subsection{STN configurations}
\begin{table}[ht!]
\begin{center}
\begin{tabu}{|c|c|c|}
\hline
\bf{Layers} & \bf{Output Size} & \bf{Configurations} \\ 
\hline\hline
ConvBlock1 & 32 $\times 50 \times 50$ & $3\times3, BN, ReLU, 2\times2$  \\   \cline{1-3}
ConvBlock2 & 64 $\times 25 \times 25$ & $3\times3, BN, ReLU, 2\times2$ \\ \cline{1-3}
ConvBlock3 & 32 $\times 12 \times 12$ & $3\times3, BN, ReLU, 2\times2$ \\ \cline{1-3}
ConvBlock4 & 16 $\times 6 \times 6$ & $3\times3, BN, ReLU, 2\times2$ \\ \cline{1-3}
FC1 & 512  &  $BN, ReLU$ \\ \cline{1-3}
FC2 & $3 \times M + 5$ & None\\ \cline{1-3}
\hline
\end{tabu}
\end{center}
\caption{The configurations of STN are shown in this table. The configuration values are arranged sequentially as follows: kernel size, normalization layer, activation layer, and max-pooling. $M$ refers to the number of line segments, which was set to $10$.}
\label{table:STN_CONFIG}
\end{table}
Table~\ref{table:STN_CONFIG} shows the configurations for the localization network within the STN. While the STN takes an input image with a resolution of $300\times300$ pixels, the localization network requires an image with a resolution of $100\times100$ pixels. This implementation is faithful to ASTER STN \cite{shi2018aster}, which also uses down-sampled images for control points prediction.

However, unlike ASTER STN, the localization network here does not output the control points directly. Instead, it predicts 5 polynomial coefficients, as well as $\hat{\phi}$, $\hat{\theta}$, and $\hat{\xi}$ for each of the $10$ line segments. Therefore, altogether the localization network's output size is $35$ for each image, and these $35$ values are converted into $30$ control points.

\subsection{Discussion on control points}
\begin{figure}[ht]
\begin{center}
   \includegraphics[width=0.6\linewidth]{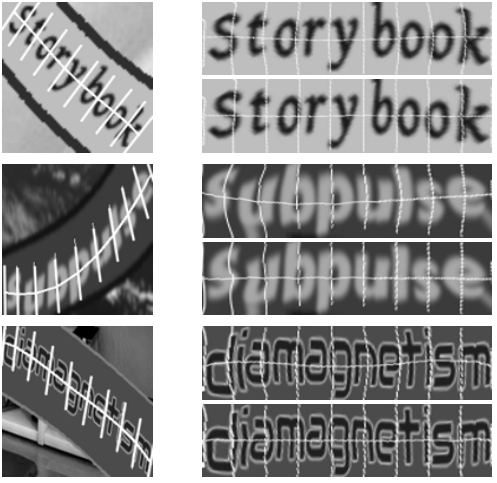}
\end{center}
\caption{Rectification results of STN using only the line segment endpoints (top) vs.  endpoints + midpoint (bottom).} 
\label{fig:STN_CP}
\end{figure}
Figure~\ref{fig:STN_CP} shows that the rectified images using three control points can better mitigate character-level distortions, and many curved characters from before are straightened when the midpoints are introduced. Moreover with the midpoints, the rectified polynomial forms a straight line across the image, which better aligns with the expected rectified outcome. Therefore, in this work, the use of three control points per line segment is favored, in contrast with similar polynomial scheme \cite{zhan2019esir} which uses only the line segments' endpoints. 

Do note that control points predicted outside the image are clamped to the image boundary as per the implementation of ASTER STN, and in that case, the truncated line segment may cause some distortions in the rectified image.

\begin{table*}[ht!]
\begin{center}
\begin{tabular}{|c|c|c|c|c|c|c|c|}
\hline
\multirow{3}{*}{\bf{Method}} & \multicolumn{3}{c|}{\bf{Regular Text}} & \multicolumn{3}{c|}{\bf{Irregular Text}} & \bf{Avg.} \\ \cline{2-7}
& IIIT & IC13 & SVT & IC15 & SVTP & CT80 & \bf{Acc}\\ \cline{2-8}
& 3000 & 1015 & 647 & 2077 & 645 & 288 & 7672\\
\hline\hline
DaViT & 94.3 & 95.7& 90.9& 77.0 & 83.4 & 84.7& 88.2\\

SaViT & 
94.3& 
95.6& 
90.4& 
76.5& 
83.9& 
81.3&
88.0$_{(\textcolor{red}{-0.2})}$\\ 
\hline
STN+DaViT & 
95.1& 93.9& 92.1& 79.6 & 87.9& 86.8 & 89.6 \\

STN+SaViT & 
94.5& 
94.0& 
92.4& 
78.5& 
87.4& 
86.8&
89.0$_{(\textcolor{red}{-0.6})}$\\ 
\cline{1-8}
\end{tabular}
\end{center}
\caption{Result of ablation study for DaViT, in terms of text recognition accuracy (in percentages). The subscript in the cell represents its relative accuracy with respect to its dual-axes counterpart.}
\label{tab:ablation-DaViT}
\end{table*}

\subsection{STN training dataset}
\begin{figure}[ht!]
\begin{center}
   \includegraphics[width=0.6\linewidth]{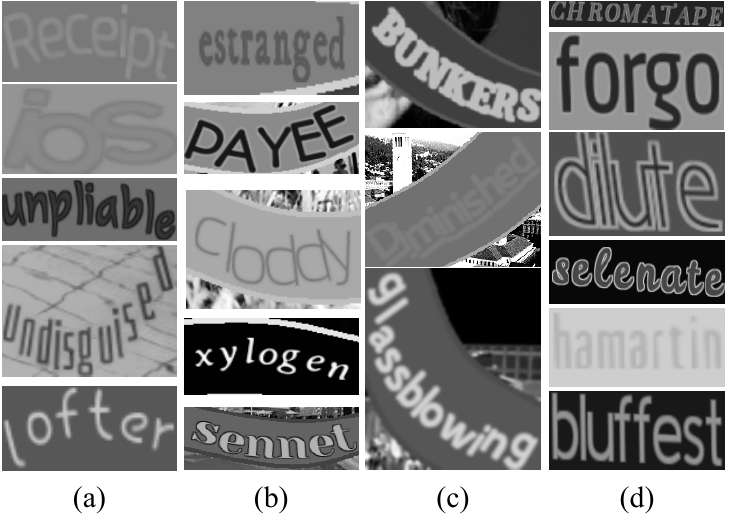}
\end{center}
\caption{Examples of the 4 types of training data for the STN: (a) curved texts, (b) curved texts with background stripes, (c) curved texts with background stripes \& image rotation, and lastly, (d) simple straight text.} 
\label{fig:STN_DS}
\end{figure}
The STN training set containing four types of images shown in Figure \ref{fig:STN_DS} was generated by the SRNet framework \cite{wu2019editing}. Customization were made to add a single band of solid-color stripe as the image background. In this paper, $20,000$ samples were generated for each type of training data.

\section{Discussion on DaViT}

In order to study the effectiveness of DaViT, the y-encoder layers were removed and the resulting model is denoted as SaViT. For the training of SaViT, the input image was sliced into $N_x \times P_wH$ strips (as opposed to patches of $P_wP_h$), where $P_w=2$ and $N_x=64$. The attention for SaViT only spans across the sequence of strips through the x-encoder layers.

Table~\ref{tab:ablation-DaViT} shows the result of the experiments where DaViT generally enhances the overall performance as compared to SaViT with a significant improvement in the irregular test datasets. This suggests that the attention in both axes allows the network to learn a feature representation that is stronger against distortions. 

\section{Examples from test datasets}
\begin{figure*}[!ht]
\begin{center}
    \includegraphics[width=0.8\linewidth]{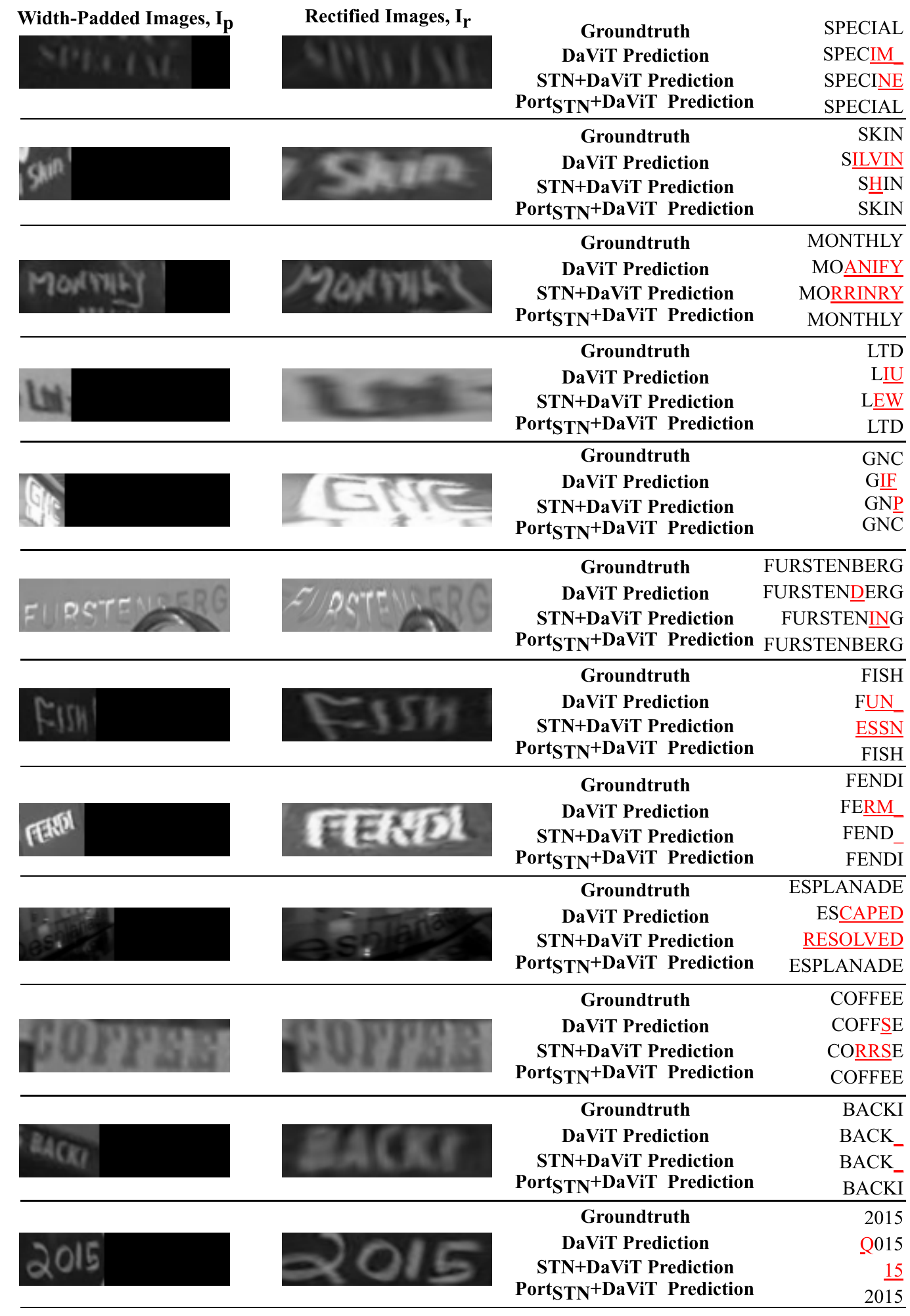}
\end{center}
\caption{Examples showing that the combination of failure cases from both DaViT and STN+DaViT, eventually produces accurate predictions when Port$_{STN}$+DaViT was applied instead.} 
\label{fig:failure1}
\end{figure*}

\begin{figure*}[!ht]
\begin{center}
    \includegraphics[width=0.85\linewidth]{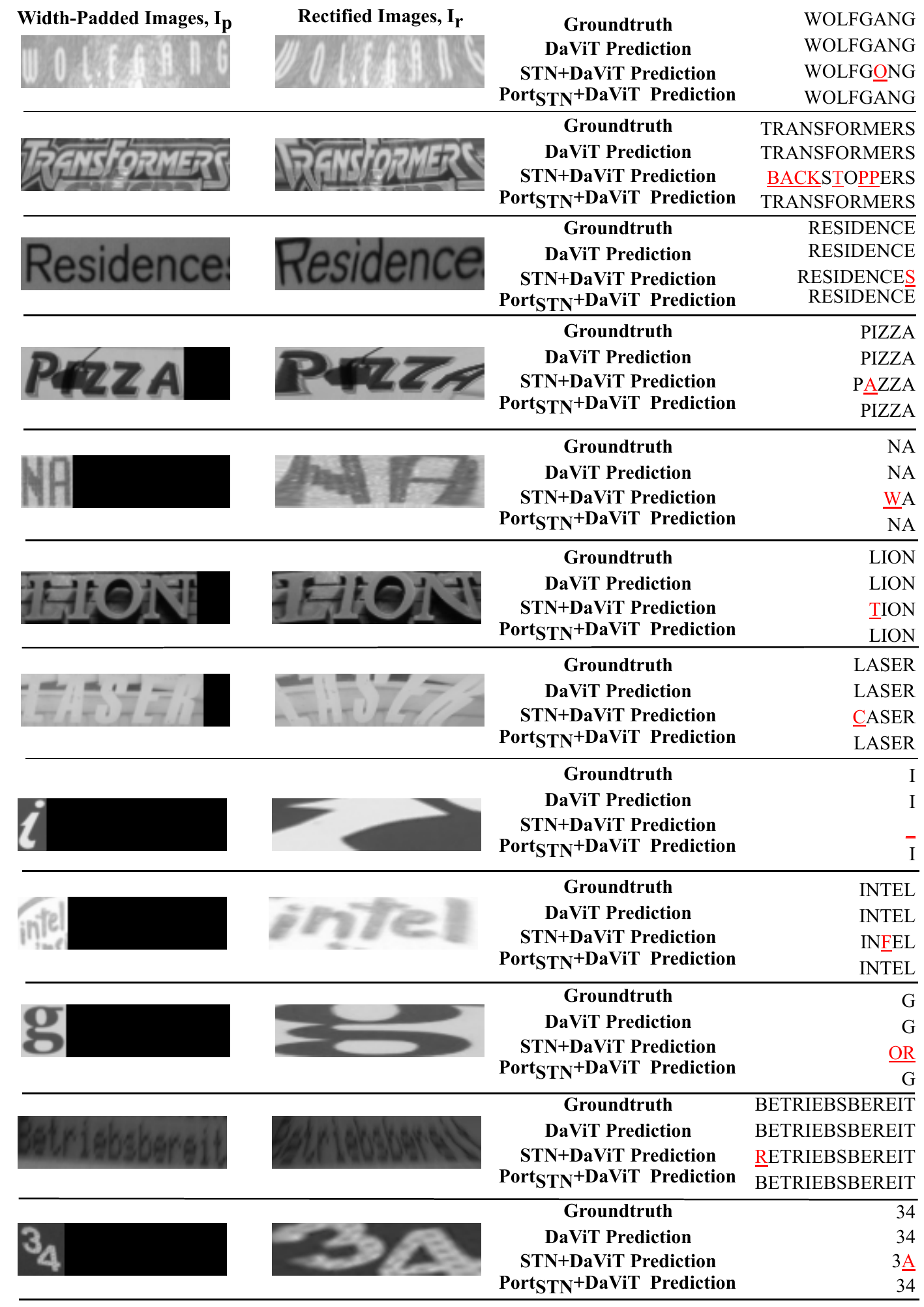}
\end{center}
\caption{Examples where correct predictions was achieved by Port$_{STN}$+DaViT mitigating the adverse impacts of rectification.} 
\label{fig:failure2}
\end{figure*}

\begin{figure*}[!ht]
\begin{center}
    \includegraphics[width=0.85\linewidth]{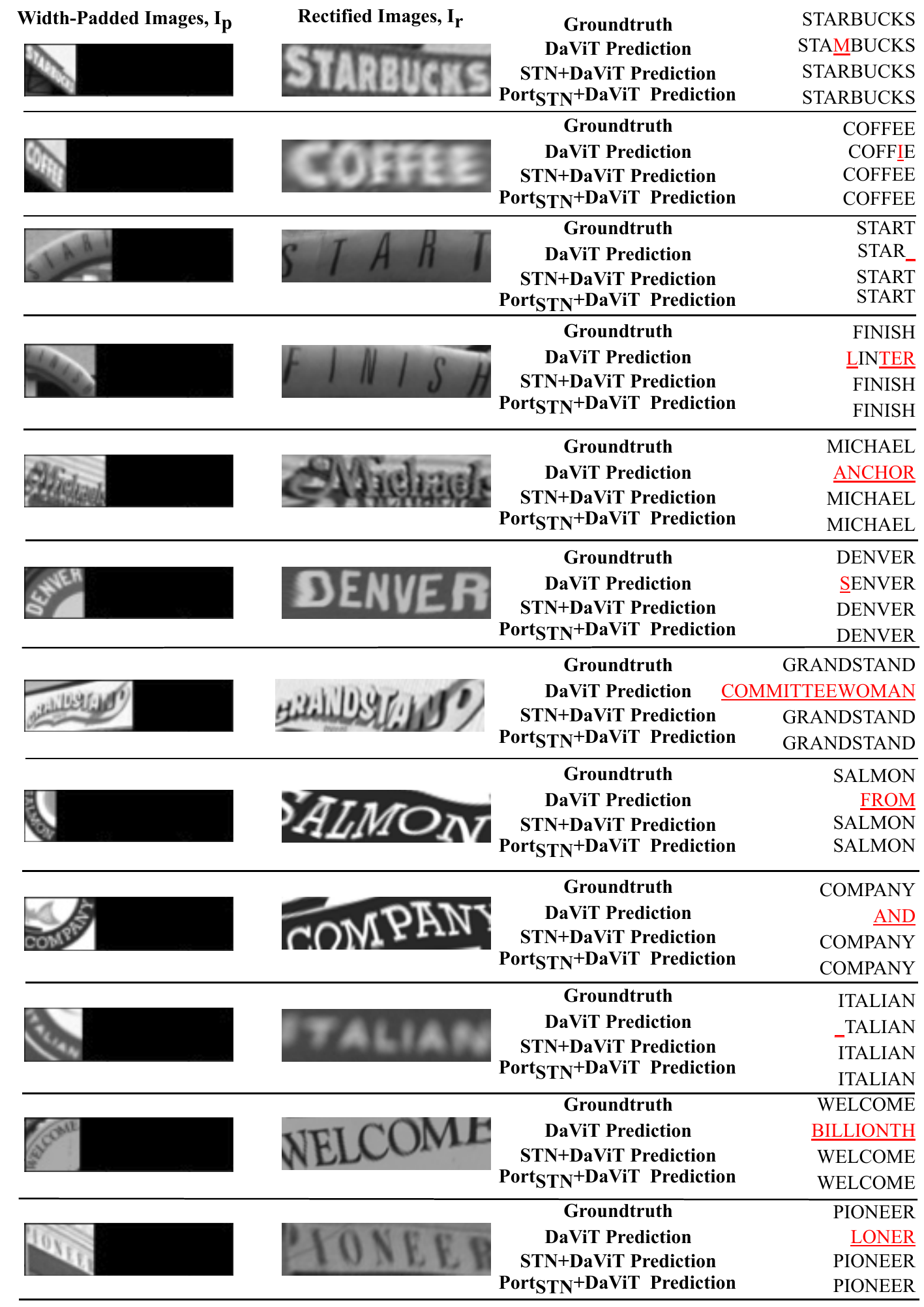}
\end{center}
\caption{Examples showing that portmanteau features did not prevent the model from leveraging the benefits of rectification networks.} 
\label{fig:failure3}
\end{figure*}

Figures \ref{fig:failure1},\ref{fig:failure2},\ref{fig:failure3} show the examples of predictions results from the models.
\label{app:examples}

\newpage






\bibliographystyle{IEEEtran}
\bibliography{IEEEabrv,ref}
%



